\documentclass[letterpaper, 10 pt, conference]{ieeeconf}  % Comment this line out if you need a4paper
\usepackage{graphicx}
\usepackage{booktabs}
\usepackage{times}
\usepackage{soul}
\usepackage[table,xcdraw]{xcolor}
\usepackage{url}
\usepackage{amsfonts,amssymb}
\usepackage[hidelinks]{hyperref}
\usepackage[utf8]{inputenc}
\usepackage[small]{caption}
\usepackage{graphicx}
\usepackage{amsmath}

\usepackage{amsthm}
\usepackage{booktabs}
\usepackage{algorithm}
\usepackage{algorithmic}
\usepackage[switch]{lineno}
\usepackage{booktabs}
 \usepackage{multirow}
 \usepackage{pifont}
 \usepackage{bm}
 \usepackage{tabularx}
 \usepackage{adjustbox}
 \usepackage{makecell}
\usepackage{array}  
\usepackage{caption}

\newcommand{\figref}[1]{Fig.~\ref{#1}}

\definecolor{iccvblue}{rgb}{0.21,0.49,0.74}
\definecolor{Light}{rgb}{0.99, 0.92, 0.95}
\usepackage{multirow}

\usepackage{tabularx}  % 自动调整列宽

\IEEEoverridecommandlockouts                              % This command is only needed if 
\title{\LARGE \bf
Multi-View Reconstruction with Global Context \\ for 3D Anomaly Detection*
}

\author{Yihan Sun$^{1}$,~\IEEEmembership{Graduate Student Member,~IEEE}, Yuqi Cheng$^{1}$,~\IEEEmembership{Graduate Student Member,~IEEE}, \\Yunkang Cao$^{1}$,~\IEEEmembership{Graduate Student Member,~IEEE}, Yuxin Zhang$^{1}$,~\IEEEmembership{Graduate Student Member,~IEEE}, \\Weiming Shen$^{1}$,~\IEEEmembership{Fellow,~IEEE}% <-this % stops a space
\thanks{*This work was partially supported by the High Quality Special Project of the Ministry of Industry and Information Technology of the People's Republic of China under Grant \#2023ZY01089 (\emph{Corresponding author: Weiming Shen}).}
\thanks{$^{1}$Yihan Sun, Yuqi Cheng, Yunkang Cao, Yuxin Zhang, and Weiming Shen are with the State Key Laboratory of Intelligent Manufacturing Equipment and Technology, Huazhong University of Science and Technology, Wuhan 430074, China.(email: yihansun@hust.edu.cn; yuqicheng@hust.edu.cn; caoyunkang@ieee.org; zyx\_hust@hust.edu.cn; wshen@ieee.org).}%
}

\begin{document}

\def\ourmethod{MVR}

\maketitle
\thispagestyle{empty}
\pagestyle{empty}

%%%%%%%%%%%%%%%%%%%%%%%%%%%%%%%%%%%%%%%%%%%%%%%%%%%%%%%%%%%%%%%%%%%%%%%%%%%%%%%%
\begin{abstract}

3D anomaly detection is critical in industrial quality inspection. While existing methods achieve notable progress, their performance degrades in high-precision 3D anomaly detection due to insufficient global information. To address this, we propose Multi-View Reconstruction (MVR), a method that losslessly converts high-resolution point clouds into multi-view images and employs a reconstruction-based anomaly detection framework to enhance global information learning. Extensive experiments demonstrate the effectiveness of MVR, achieving 89.6\% object-wise AU-ROC and 95.7\% point-wise AU-ROC on the Real3D-AD benchmark. The code is available at: \url{https://github.com/hustSYH/MVR}

\end{abstract}

%%%%%%%%%%%%%%%%%%%%%%%%%%%%%%%%%%%%%%%%%%%%%%%%%%%%%%%%%%%%%%%%%%%%%%%%%%%%%%%%

\section{Introduction}

3D anomaly detection technology provides innovative solutions for defect identification through high-precision 3D scanning~\cite{Survey}, with extensive applications in precision manufacturing quality inspection~\cite{MVGR}. Its core objective is to detect abnormal points that deviate from normal patterns for 3D structures. Unsupervised 3D anomaly detection has emerged as the predominant industrial inspection paradigm~\cite{cpmf,MiniShift_Simple3D}, requiring only anomaly-free training samples while eliminating dependency on annotated defective data. However, the performance of state-of-the-art (SOTA) 3D anomaly detection methods remains insufficient for industrial deployment due to two fundamental limitations: i) limited feature learning capacity leading to loss of essential anomaly-related details during extraction, and ii) insufficient utilization of global contextual information in post-extraction processing. This study focuses on advancing 3D anomaly detection by developing a multi-view reconstruction framework with global context to meet rigorous real-world performance requirements.

\begin{figure}[htbp]
\centering
\includegraphics[width=1\linewidth]{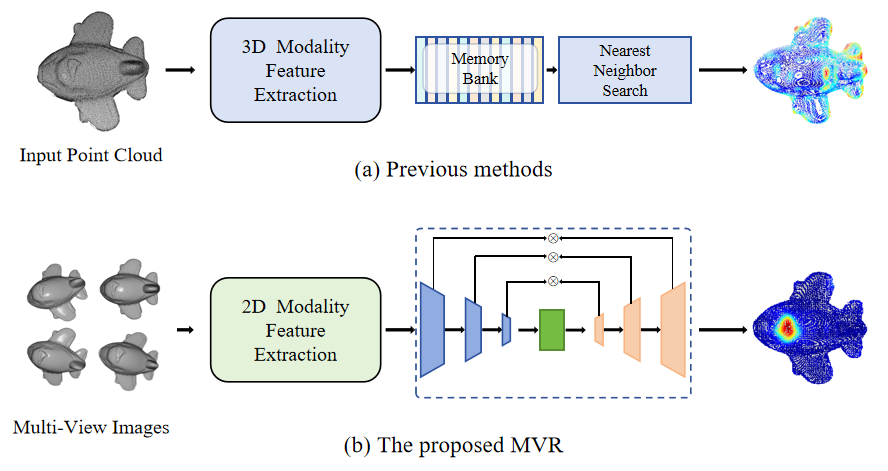}
\caption{Different from previous methods and the proposed MVR.}
\vspace{-5mm}
\label{fig:fig1}
\end{figure}

Global features comprise the overall structure and statistical contextual correlations of point clouds and thus can effectively capture cross-regional dependencies and avoid interference from local noise, making them crucial in 3D anomaly detection. However, existing SOTA methods tend to prioritize local semantics, demonstrating notable deficiencies in extracting and leveraging global semantics. Reg3D-AD~\cite{read3d}, a registration-based 3D anomaly detection method, primarily relies on local coordinate-based features and thus loses sight of global features. Group3AD~\cite{Group3d} extracts group-level features from the input point cloud by minimizing the intra-group distance and maximizing the inter-group distance. However, such a confounding feature representation also brings about negative impacts. GLFM~\cite{Cheng2025BoostingGF} proposes the use of pre-trained point transformers to capture both global and local contextual information, and specifically performs clustering on global semantics. However, its design based on the memory bank overly relies on local features and fails to make good use of the extracted global features.

To enhance feature extraction and global contextual understanding, we propose  Multi-View Reconstruction (MVR), which firstly converts unstructured point clouds into high-fidelity structured image representations. This transformation enables the utilization of pre-trained Vision Transformers (ViTs) to effectively model long-range dependencies through self-attention mechanisms, thereby capturing comprehensive global semantic information.
While CPMF~\cite{cpmf} introduces a multi-view rendering approach, our method distinguishes itself through a high-resolution projection strategy that explicitly addresses rendering artifacts and quality degradation, as detailed in Section~\ref{sec:abaltions}. Furthermore, unlike memory-bank-based approaches, MVR adopts a reconstruction-driven framework that enforces implicit global constraints during training. This paradigm shift facilitates the recovery of spatial structure while promoting the learning of globally coherent semantic representations. 
As depicted in Fig.~\ref{fig:fig1}, MVR integrates a reconstruction-based 2D feature mapping mechanism with multi-view rendering. This design contrasts with conventional 3D feature extraction methods that rely on memory-bank-based local matching. By holistically capturing global semantic correlations, MVR achieves SOTA performance in 3D anomaly detection.
Extensive experiments conducted on the high-resolution Real3D-AD benchmark~\cite{read3d} demonstrate the efficacy of MVR. Results indicate that MVR achieves object-wise and point-wise AUROC scores of 89.6\% and 95.7\%, respectively, significantly outperforming previous state-of-the-art methods. Our contributions are summarized as follows:

\begin{itemize}
\item A multi-view rendering strategy that losslessly transforms high-resolution point clouds into multi-view images while preserving structural integrity.
\item A reconstruction-based paradigm that explicitly enforces global context learning through spatial structure recovery, enhancing semantic representation quality.
\end{itemize}

\begin{figure*}[htbp]
\centering
\includegraphics[width=0.9\linewidth]{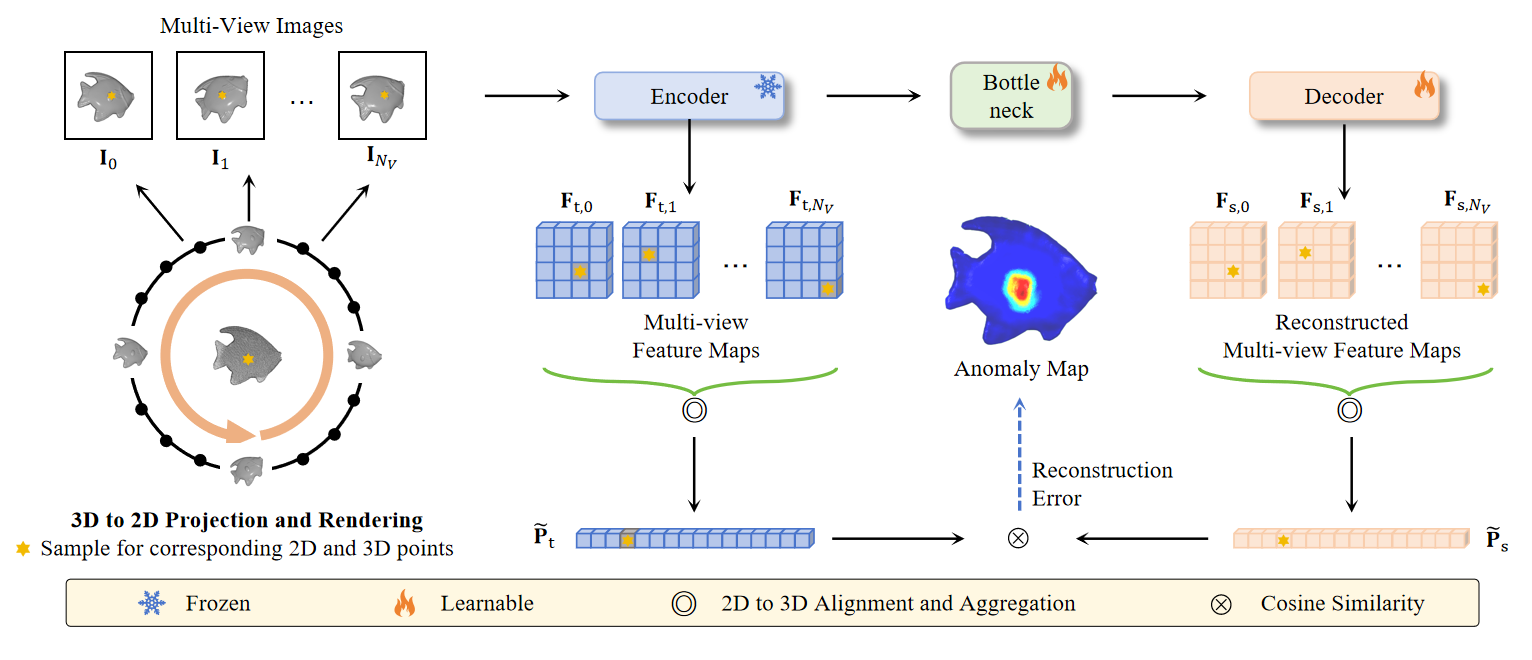}
\caption{Framework of the proposed method.}
\vspace{-3mm}
\label{fig:fig2}
\end{figure*}

\section{MVR Method}

\subsection{Problem Definition} Given an input point cloud 
$\mathbf{P} \in \mathbb{R}^{N\times3}$ (containing N spatial points), the goal of 3D anomaly detection is to learn a mapping function $\Phi: \mathbf{P} \rightarrow(\xi,\left\{s_{i}\right\}_{i=1}^{N} )$, which assigns the object-wise anomaly score $\xi \in [0, 1]$ and the point-wise anomaly score $s_{i} \in \mathbb{R}^+$ , thereby constructing a point-wise anomaly map $\mathbf{S}=\left\{s_{i}\right\}_{i=1}^{N}$. A higher $s_{i}$ indicates a higher probability that $p_{i}$ belongs to an anomalous region. Typically, the model is trained exclusively on normal samples $\mathbf{P}^{train} \sim \mathcal{D}_{\text {normal }}$, while Testing under $\mathbf{P}^{test} \sim \mathcal{D}_{\text {normal }}\cup\mathcal{D}_{\text {anomaly }} $ with open-set anomalies.

\subsection{Overview} Existing 3D anomaly detection approaches are predominantly based on memory bank mechanisms, emphasizing localized feature extraction at the expense of comprehensive global context modeling, which consequently results in suboptimal detection accuracy. To address this limitation and holistically leverage global information in point clouds, this study proposes a novel multi-view reconstruction framework, shown in \figref{fig:fig2}.

It consists of two principal components: 1) A multi-view projection architecture that converts irregular point clouds into structured depth map representations, and 2) A hierarchical reconstruction network comprising three core components: a multi-scale encoder for feature abstraction, a latent-space bottleneck for information compression, and a multi-stage decoder for geometric reconstruction.

The hierarchical reconstruction network systematically processes projected depth maps through feature embedding and reconstruction stages. Based on view-point cloud geometric correspondence, it implements cross-view feature coordination before and after depth map reconstruction, then transfers multi-view features back to the original point cloud space via inverse projection mapping. The constructed fused point cloud feature representation facilitates model training and residual-based anomaly detection.

\subsection{Multi-View Projection}

To leverage mature pre-trained 2D neural networks for feature extraction, this study employs a 3D to 2D projection and rendering approach to process point clouds, transforming unstructured point clouds into high-quality structured images. The input point cloud is projected onto multiple predefined viewpoints, generating a set of 2D view images 
$\mathbf{I}$
$( {\mathbf{I}_1,\mathbf{I}_2, ..., \mathbf{I}}_n \in \mathbb{R}^{H \times W \times 3} )$, where $ H \times W $ denotes the spatial resolution of rendered images and n is the number of rendering views.

 Let $\mathbf{P}^i$ be the 3D position of the \textit{i}-th point, and $\mathbf{I}^i$ be the corresponding point in the pixel coordinate system.The projection relationship between $\mathbf{P}^i$ and $\mathbf{I}^i$ is given by:

\begin{equation}
\left[ \mathbf{I}^{i}, 1 \right]^{\mathrm{T}} = \frac{1}{z_{i}} \mathbf{K} \mathbf{T} \left[ \mathbf{P}^{i}, 1 \right]^{\mathrm{T}} 
\end{equation}
where $\mathbf{T}$ is the homogeneous transformation matrix from the point cloud coordinate system $\left\{{\mathbf{T}_p}\right\}$ to the camera coordinate system $\left\{{\mathbf{T}_c}\right\}$,$z_{i}$ is the Z-coordinate of $\mathbf{P}^i$ in $\left\{{\mathbf{T}_c}\right\}$,and $\mathbf{K}$ is the intrinsic matrix of the camera.

To enhance the representational completeness of data characterization and mitigate the inherent limitations of single-view observations, this study employs sequential rigid transformations using a set of distinct rotation matrices for iterative spatial coordinate transformations. Setting different poses $ \mathbf{T}_k$,$1\le k \le N_v$,multiple depth images $ \mathbf{I}_k$,$1\le k \le N$ can be generated after projection and rendering, as shown in Fig. 3. Let $ \mathbf{I}_k^i$ be the corresponding  point of $\mathbf{P}^i$ in $ \mathbf{I}_k$, and the projection relationship between $\mathbf{P}^i$ and $ \mathbf{I}_k^i$ is given by:

\begin{equation}
\left[ \mathbf{I}_k^{i}, 1 \right]^{\mathrm{T}} = \frac{1}{z_{i}} \mathbf{K} \mathbf{T}_k \left[ \mathbf{P}^{i}, 1 \right]^{\mathrm{T}} 
\end{equation}

Define $\operatorname{Pos}_{\mathrm{k}}: \mathbf{I}_k \longmapsto \mathbf{P}$ as the mapping from image pixels to points in the point cloud:

\begin{equation}
\mathbf{P}=\operatorname{Pos}_{\mathrm{k}}\left(\mathbf{I}_{k}\right)=\mathbf{R}_{k}^{-1}\left(z_{i} \mathbf{K}^{-1} \mathbf{I}_{k}^{i}-\mathbf{t}_{k}\right) 
\end{equation}
where $\mathbf{R}_{k}$ is the rotation matrix of $\mathbf{T}_k$, and $\mathbf{t}_{k}$ is the translation vector of $\mathbf{T}_k$. $\operatorname{Pos}_{\mathrm{k}}(\cdot)$ function aligns the features of points in the subsequent integration process.

\subsection{MVR}
With multi-view rendering, images from multiple views are acquired. Then, this study employs Dinomaly~\cite{dinomaly} —a state-of-the-art 2D reconstruction method, to capture global information more comprehensively. Dinomaly employs scalable foundation transformers that extracts universal and discriminative features.  Furthermore, it leverages the ”side effect” of Linear Attention that impedes focus on local regions, thus promotes focus on global information. Building upon the above design, Dinomaly suggests using the reconstruction errors between a teacher and a student network to score anomalies, with the underlying assumption that by training with solely normal samples $\mathbf{P}^{train}$.

\subsubsection{ Pre-trained Feature Extraction }
This study leverages the global receptive field generated by Vision Transformer (ViT)'s self-attention mechanism to extract global features. The encoded feature representation for each depth map is subsequently derived through a pre-trained ViT encoder~\cite{Luo2025ExploringIN} (comprising 12 transformer layers), which implicitly learns the data manifold structure to map normal samples of varying sizes to distinct positions in the latent space while preserving their topological consistency. This process can be formally expressed as: 

\begin{equation}
\mathbf{F}_t^{(k)} = \mathcal{E}(\mathbf{I}_k) , \quad \forall k \in {1, 2, ..., N_v} 
\end{equation}
where $ \mathbf{F}_t^{(k)}$ is the feature map of the teacher network in the \textit{k}-th view.

Thus, we can extract j intermediate-layer feature maps $\mathbf{F}_t^{(k)}= \left\{ \mathbf{f}_{t,1}^{(k)}, \ldots,\mathbf{f}_{t,j}^{(k)} \right\}$, which constitute the teacher feature set of the \textit{k}-th view.

\subsubsection{ Multi-View Feature Reconstruction } 
The bottleneck is a MLP that collects the feature representations of the encoder’s j middle-level layers. The decoder is similar to the encoder, comprising j Transformer layers, can automatically adjust to size changes and supplement global features during reconstruction from latent variables, ultimately rebuilding the student feature maps $\mathbf{F}_s^{(k)}= \left\{ \mathbf{f}_{s,1}^{(k)}, \ldots,\mathbf{f}_{s,j}^{(k)} \right\}$.

Then, point-wise feature maps from different transformer layers are aggregated with an average pooling.
\begin{equation}
\mathbf{F}_{t,k}=\frac{1}{j} \sum_{l=1}^{j} \mathbf{f}_{t,l}^{(k)} 
\end{equation}
\begin{equation}
\mathbf{F}_{s,k}=\frac{1}{j} \sum_{l=1}^{j} \mathbf{f}_{s,l}^{(k)} 
\end{equation}
where $\mathbf{F}_{t,k}$ and $\mathbf{F}_{s,k}$ both are the aggregated point-wise 2D modality feature maps composing cues from different transformer layers.

\subsubsection{ Multi-View Feature Fusion } 
Benefiting from the prior knowledge learned by the pre-trained 2D neural network on large-scale image datasets and the strong representational capacity of the reconstruction method for global features, both the teacher feature maps $\mathbf{F}_{t,k}$ and student feature maps $\mathbf{F}_{s,k}$ exhibit more distinctive and comprehensive information. The next step is to aggregate the features of each point from the multi-view depth images. The point features $\mathbf{P}_{t,k}^i$ and $\mathbf{P}_{s,k}^i$ of point $\mathbf{P}^i$ on the \textit{k}-th depth image can be obtained as:

\begin{equation}
\mathbf{P}_{t,k}^i=\operatorname{Pos}_{\mathrm{k}}\left(\mathbf{F}_{t,k}^{i}\right) 
\end{equation}
\begin{equation}
\mathbf{P}_{s,k}^i=\operatorname{Pos}_{\mathrm{k}}\left(\mathbf{F}_{s,k}^{i}\right) 
\end{equation}
Define $\tilde{\mathbf{P}}$ as the point features aggregated from multi-view depth images, and the \textit{i}-th point features $\tilde{\mathbf{P}}_t^i$ and $\tilde{\mathbf{P}}_s^i$ can be obtained as:
\begin{equation}
\tilde{\mathbf{P}}_t^i=\frac{1}{N} \sum_{k}^{N} \mathbf{P}_{t,k}^{i} 
\end{equation}
\begin{equation}
\tilde{\mathbf{P}}_s^i=\frac{1}{N} \sum_{k}^{N} \mathbf{P}_{s,k}^{i} 
\end{equation}

In summary, denoting the function for extracting 2D modality features as $\mathcal{E}_{ts}$, the extraction of $\tilde{\mathbf{P}}_t$ and $\tilde{\mathbf{P}}_s$ is simplified as:
\begin{equation}
\tilde{\mathbf{P}}_t,\tilde{\mathbf{P}}_s=\mathcal{E}_{ts}(\textbf{P}) 
\end{equation}
and \(\mathcal{E}_{ts}\) is utilized to detect anomalies, $\textbf{P}$ is the input point cloud.

\subsubsection{ Training } 
During the training process,we employ the hard-mining global cosine loss~\cite{Guo2023ReContrastDA} that detaches the gradients of well-restored feature points with low cosine distance. The loss function $\mathcal{L}$ is:
\begin{equation}
\mathcal{L}_{\text{global-hm}} = \mathcal{D}_{\text{cos}}(\mathcal{F}(\tilde{\mathbf{P}}_t), \mathcal{F}({\tilde{\mathbf{P}}_s}^*)) 
\end{equation}

\begin{equation}
{\tilde{\mathbf{P}}_s}^*(i) = \left\{ 
\begin{array}{ll}
sg({{\mathbf{P}}_s}^*(i))_{0.1}, \text{if } \mathcal{D}_{\text{cos}}(\tilde{\mathbf{P}}_t, \tilde{\mathbf{P}}_s) < k\%_\text{batch} \\
\tilde{\mathbf{P}}_s(i), \text{else}
\end{array}
\right. 
\end{equation}

\begin{equation}
\mathcal{D}_{\text{cos}}(a, b) = 1 - \frac{a^{\mathrm{T}} \cdot b}{\|a\| \|b\|} 
\end{equation}
where $\mathcal{D}_{\text{cos}}(\cdot,\cdot)$ denotes cosine distance, $ \mathcal{F}(\cdot)$ denotes flatten operation, $\tilde{\mathbf{P}}_s(i)$ represents the feature for the i-th point, and $sg(\cdot)_{0.1}$ denotes to shrink the gradient to one-tenth of the original. $\mathcal{D}_{\text{cos}}(\tilde{\mathbf{P}}_t, \tilde{\mathbf{P}}_s) < k\%_\text{batch}$ selects k\% feature points with smaller cosine distance specifically within a batch for gradient shrinking.

\subsection{3D Anomaly Detection} In the testing phase, to infer anomaly scores for a testing point cloud $\mathbf{P}^{test}$, this study extracts 2D modality features for $\mathbf{P}^{test}$ and then calculate anomaly scores via cosine similarity metric. The anomaly score calculation process can be formulated as follows,

\begin{equation}
\tilde{\mathbf{P}}_t^{test},\tilde{\mathbf{P}}_s^{test}=\mathcal{E}_{ts}(\mathbf{P}^{test}) 
\end{equation}

\begin{equation}
\mathbf{S}_{\text{anomaly}}^{(i)} = 1 - \frac{\tilde{\mathbf{P}}_t^{test(i)} \cdot \tilde{\mathbf{P}}_s^{test(i)}}{\|\tilde{\mathbf{P}}_t^{test(i)}\| \|\tilde{\mathbf{P}}_s^{test(i)}\|} 
\end{equation}

\begin{equation}
\xi=max(\mathbf{S}_{\text{anomaly}}^{(i)}) 
\end{equation}
where $\mathbf{S}_{\text{anomaly}}^{(i)}$ refers to the anomaly score for the \textit{i}-th point, $\mathbf{S}_{\text{anomaly}}$ is the point-wise anomaly map and $\xi$ is the object-wise anomaly score.

\section{Experiments}

\begin{table*}[t]
\centering
\caption{\textbf{Quantitative Results on Real3D-AD~\cite{read3d}}. The results are presented in O-ROC\%/P-ROC\%. The best performance is in \textbf{bold}, and the second best is \underline{underlined}. M3DM-PM and M3DM-PB for M3DM-PointMAE and M3DM-PointBERT, respectively.}
\label{table:real}
\setlength{\tabcolsep}{11pt}
\fontsize{11}{14}\selectfont{
\resizebox{0.95\linewidth}{!}{
\begin{tabular}{c|ccccccccc|
>{\columncolor{blue!8}}c}
\toprule[1.5pt]
Method~$\rightarrow$    & \small{M3DM-PM~\cite{M3DM}} & \small{M3DM-PB~\cite{M3DM}} & \small{R3D-AD~\cite{r3d}}  & \small{CPMF~\cite{cpmf}} & 
\small{Reg3D-AD~\cite{read3d}} & \small{Group3AD~\cite{Group3d}}   & \small{IMRNet~\cite{shape_anomaly}}    & \small{PointCore~\cite{pointcore}}    & \small{GLFM~\cite{Cheng2025BoostingGF}}       &\textbf{MVR}       \\
Category~$\downarrow$ & \small{CVPR'23}      & \small{CVPR'23}  &  \small{ECCV'24}   & \small{PR'24} & \small{NeurIPS'23} & \small{ACM MM'24}   & \small{CVPR'24} & \small{Arxiv'24}  & \small{TASE'25} & \textbf{Ours}      \\ \midrule
Airplane                   & 43.4/53.0                                                  & 40.7/52.3                                                  & \textbf{77.2}/-                                                   & 63.2/61.8                 & 71.6/63.1                                                     & 74.4/63.6                                                      & \underline{76.2}/-                                       & 66.0/60.8           & 73.3/\textbf{86.3}               & 67.0/\underline{84.3}     \\
Car                        & 54.1/60.7                                                  & 50.6/59.3                                                  & 69.6/-                                                   & 51.8/83.6                          & 69.7/71.8                                                   & 72.8/74.5                                                      & 71.1/-                                           &  \underline{86.6}/70.6            & 53.4/\underline{92.7}          & \textbf{87.7}/\textbf{97.4}     \\
Candybar                   & 45.0/68.3                                                  & 44.2/68.2                                                  & 71.3/-                                                   & 71.8/73.4                          & 82.7/72.4                                                     & 84.7/73.8                                                      & 75.5/-                                       & \underline{97.6}/76.0            & 59.2/\underline{94.4}                & \textbf{99.8}/\textbf{99.4}     \\
Chicken                    & 68.3/73.5                                                  & 67.3/79.0                                                  & 71.4/-                                                   & 64.0/55.9                          & \underline{85.2}/67.6                                                     & 78.6/75.9                                                      & 78.0/-                              & 84.1/78.0                   & 57.4/\underline{84.9}                & \textbf{86.5}/\textbf{85.7}     \\
Diamond                    & 60.2/61.8                                                  & 62.7/59.4                                                  & 68.5/-                                                   & 64.0/75.3                          & 90.0/83.5                                                    & 93.2/86.2                                                      & 90.5/-                                         & \underline{96.3}/81.0              & 57,8/\underline{87.2}           & \textbf{100.0}/\textbf{99.4}     \\
Duck                       & 43.3/67.8                                                  & 46.6/66.8                                                  & \textbf{90.9}/-                                                   & 55.4/71.9                          & 58.4/50.3                                                     & 67.9/63.1                                                      & 51.7/-                                                   & 68.4/71.2        & 30.2/\underline{85.7}        & \underline{83.2}/\textbf{98.1}     \\
Fish                       & 54.0/60.0                                                  & 55.6/58.9                                                  & 69.2/-                                                   & 84.0/\underline{98.8}                          & 91.5/82.6                                                     & 97.6/83.6                                                      & 88.0/-                                        & \underline{99.2}/78.2          & 69.3/94.3                & \textbf{100}/\textbf{99.6}     \\
Gemstone                   & 64.4/65.4                                                  & 61.7/64.6                                                  & 66.5/-                                                   & 34.9/44.9                          & 41.7/54.5                                                     & 53.9/56.4                                                      & \underline{67.4}/-                              & 53.4/51.5                & 34.0/\underline{84.7}                   & \textbf{93.8}/\textbf{98.7}     \\
Seahorse                   & 49.5/56.1                                                  & 49.4/57.4                                                  & 72.0/-                                                   & 84.3/96.2                         & 76.2/81.7                                                     & 84.1/82.7                                                      & 60.4/-                                                 & \textbf{97.3}/84.1       & 47.5/\underline{96.8}         & \underline{93.5}/\textbf{98.2}     \\
Shell                      & 69.4/74.8                                                  & 57.7/73.2                                                  & \underline{84.0}/-                                                   & 39.3/72.5                          & 58.3/81.1                                                     & 58.5/79.8                                                      & 66.5/-                                  & \textbf{88.1}/78.1            & 25.1/\underline{83.6}                   & 73.2/\textbf{90.4}     \\
Starfish                   & 55.1/55.5                                                  & 52.8/56.3                                                  & \underline{70.1}/-                                                   & 52.6/80.0                          & 50.6/61.7                                                     & 56.2/62.5                                                      & 67.4/-                                                      & 65.2/73.6     & 56.0/\underline{92.8}      & \textbf{91.4}/\textbf{97.9}     \\
Toffees                    & 55.2/67.9                                                  & 56.2/67.7                                                  & 70.3/-                                                   & 84.5/\underline{95.9}                          & 68.5/75.9                                                     & 79.6/80.3                                                      & 77.4/-                                                  & \underline{92.9}/74.5     & 75.5/93.9          & \textbf{99.4}/\textbf{99.0}     \\ \midrule
Mean                       & 55.2/63.7                                                  & 53.8/63.6                                                  & 73.4/-                                                   & 62.5/75.9                          & 70.4/70.5                                                     & 75.1/73.5                                                      & 72.5/-                                                  & \underline{82.9}/73.1      & 53.2/\underline{89.8}         & \textbf{89.6}/\textbf{95.7}     \\ \bottomrule[1.5pt]
\end{tabular}}}
\end{table*}

\begin{figure*}[htbp]
\centering
\includegraphics[width=0.95\linewidth]{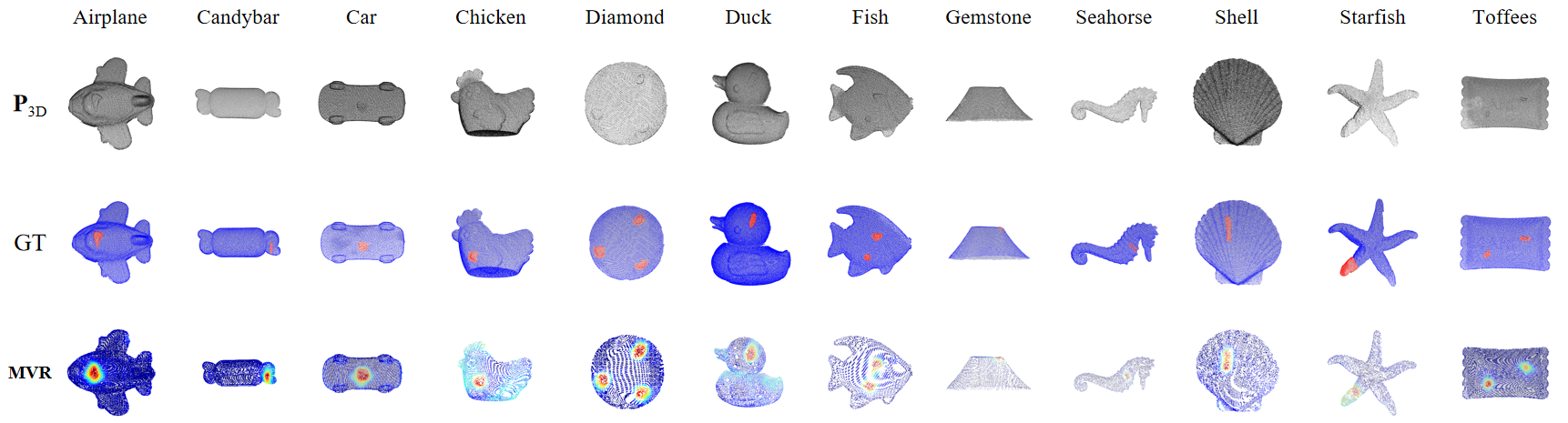}
\caption{Visualization of prediction results using the proposed method on Real3D-AD~\cite{read3d}.}
\vspace{-3mm}
\label{fig:fig3}
\end{figure*}

\subsection{Experimental Settings}

\subsubsection{ Dataset }
This study conducts experiments on the Real3D-AD dataset[1], which specifically contains high-resolution scans of 12 different object categories obtained from industrial 3D sensors, respectively. With 1,254 high-resolution 3D items (from forty thousand to millions of points for each item), Real3D-AD is a high-quality dataset for high-precision 3D industrial anomaly detection. In this study, background points are filtered out in the Real3D-AD dataset, and it is projected into multi-view depth images, with a total of 27 depth images used.

\subsubsection{ Implementation Details }
This study renders multiview images for the given point cloud using the implementations of the Open3D library~\cite{Zhou2018Open3DAM}, and the rendered images are fixed with a spatial resolution of 672 × 672 by default. After generating multi-view images at high resolution, they are resized to 224×224 as the input images. ViT-Large/14 (patchsize=14) pre-trained by DINOv2-R~\cite{DINOv2-R} is used as the encoder, with a 1024-dimensional embedding feature. In this work, we selected the feature representations of the encoder’s 8 middle-level layers, and the corresponding decoder is composed of 8 Transformer layers. The stableAdamW optimizer~\cite{Wortsman2023StableAdamL} with AMSGrad~\cite{AMSGrad} (more stable than AdamW in training) is used with lr=2e-4 and wd=1e-5. The network is trained for 300 iterations (steps) on Real3D-AD. The model inference is conducted on the GPU RTX 3090Ti, employing PyTorch version 2.6.

\subsubsection{ Evaluation Metrics }
In this study, object-wise anomaly detection is measured using object-wise ROCAUC (denoted O-ROC). For point-wise analysis, a point-wise ROCAUC metric is employed, which extends the standard ROCAUC to the pixel domain by treating each individual pixel as an independent sample and computing the area under the curve across all pixels in the dataset (denoted P-ROC).

\subsection{Comparisons with SOTA Methods}
Table 1 summarize per-class comparisons between our method and other state-of-the-art methods, including R3D-AD~\cite{r3d}, CPMF~\cite{cpmf}, Group3AD~\cite{Group3d}, IMRNet~\cite{shape_anomaly}, PointCore~\cite{pointcore} and several benchmarking methods reported in Real3D-AD~\cite{read3d}.

For O-ROC metric, while PointCore demonstrates the best average performance among existing methods, its 82.9\% O-ROC remains suboptimal. By contrast, the proposed method achieves significantly higher performance (89.6\% O-ROC), outperforming all baselines. Specifically, our method leads in eight out of twelve categories and ranks second in the remaining two, solidifying its dominance.

Regarding the P-ROC metric, our findings indicate that our method exhibits a  marked improvement over CPMF across various test scenarios.The proposed method attains an outstanding 95.7\% average P-ROC, surpassing the CPMF baseline (75.9\%) by an absolute margin of 19.4 percentage points. Notably, our approach attains the highest scores across eleven categories. These results indicate our method's effectiveness in handling complex 3D anomaly detection tasks and its distinct robustness advantage.

\figref{fig:fig3} provides a visual representation of the clustering results  achieved with our method on the Real3D-AD dataset, which clearly shows the effectiveness of our method in localizing geometrical abnormalities.

\subsection{Ablation Studies}~\label{sec:abaltions}
In this subsection, we investigate the effects of rendering resolution and the number of rendering views employed during testing phases.

\begin{figure}[htbp]
\centering
\includegraphics[width=1\linewidth]{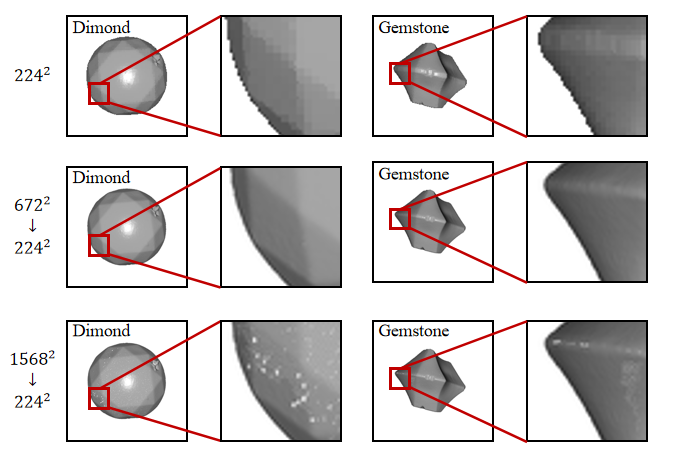}
\caption{Illustration of image quality comparison under different rendering resolutions.}
\vspace{-5mm}
\label{fig:fig4}
\end{figure}

\begin{figure}[htbp]
\centering
\includegraphics[width=1\linewidth]{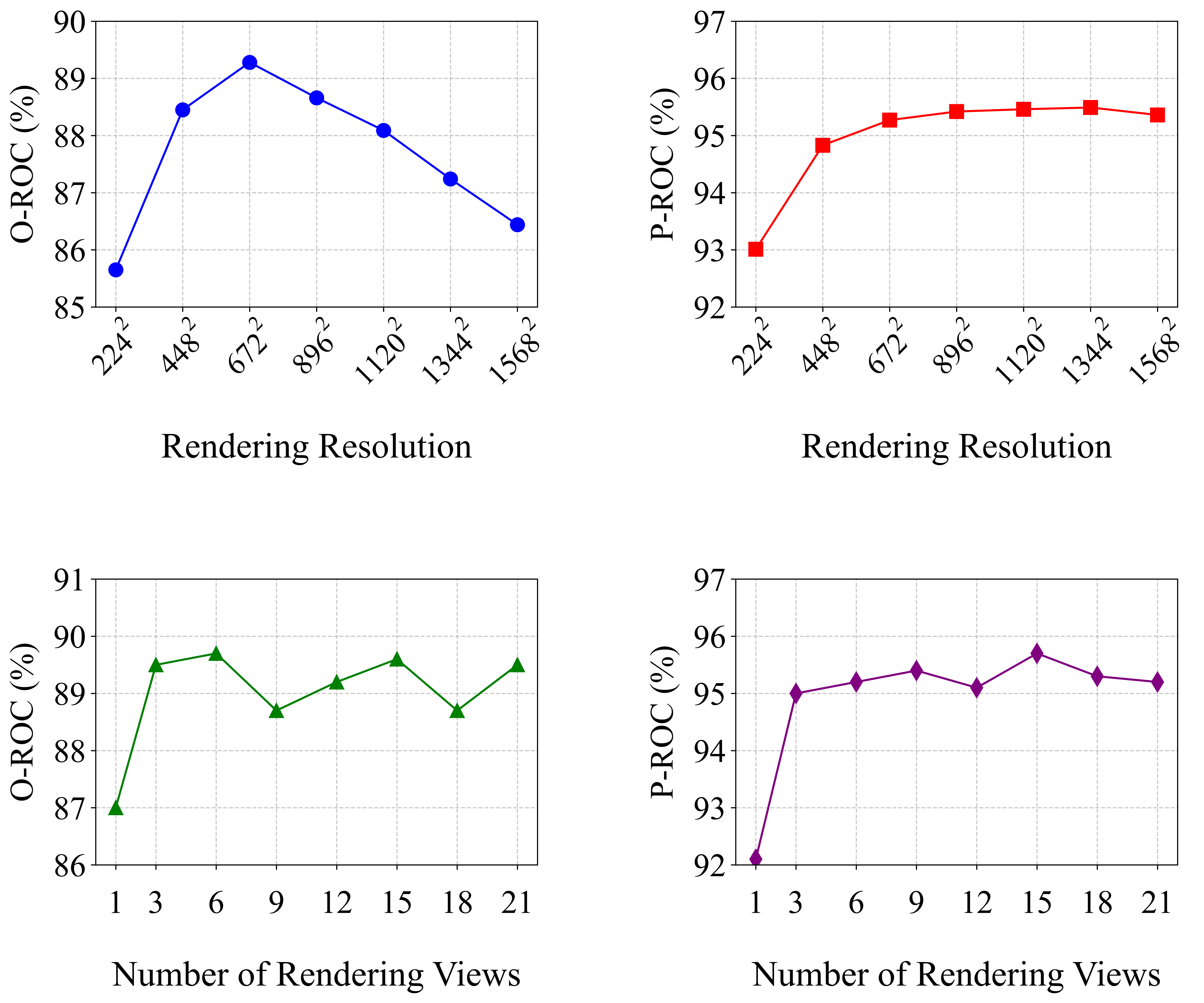}
\caption{Comparisons of the anomaly detection performances under different rendering resolutions and different number of views.}
\vspace{-5mm}
\label{fig:fig5}
\end{figure}

\subsubsection{ Influence of Rendering Resolution }
CPMF generate $224^2$ depth views via 3D to 2D projection and rendering. However, inherent noise in point clouds acquisition degrades direct rendering quality, potentially causing feature distortion and misclassification. To enhance feature discriminative capability, our method implements a high-resolution rendering strategy followed by downsampling to $224^2$. Through systematic experiments with rendering resolutions ($224^2-1568^2$), \figref{fig:fig5} reveals: O-ROC increases notably by 3.6\% as the resolution rises from $224^2$ to $672^2$, reaching its peak(89.3\%) at $672^2$, then decaying to 86.4\% at $1568^2$. P-ROC demonstrates moderate fluctuations (+2.5\% cumulative gain) across $224^2-1344^2$ resolutions before decreasing.

The overall improvements can be explicitly explained as follows: when point clouds are directly projected to low resolution, multiple 3D points may map to the same pixel, resulting in information loss and projection conflicts, especially in cases of under non-uniform point distributions where noise artifacts become more pronounced. Conversely, high-resolution rendering captures fine geometric details (e.g., edges, micro-structures) through higher pixel density, effectively mitigating aliasing and quantization errors. Subsequent downsampling to $224^2$ further helps in enabling algorithmic smoothing of high-frequency noise via pixel aggregation, implicitly implementing anti-aliasing while suppressing stochastic noise components.

For the observed decline in O-ROC metrics from $672^2$ to $1568^2$ resolution, our analysis of individual category-specific O-ROC indicators reveals non-uniform performance variations across classes: most categories exhibit reduced detection performance while a small subset shows improvement. Upon inspecting rendered multi-view projections, we identified that high-resolution depth maps generated from sparse point clouds (e.g., dimond, gemstone) develop regional data voids. These voids introduce anomalous noise artifacts during downsampling to $224^2$ resolution through interpolation processes, ultimately degrading performance by corrupting critical features with noise interference.

\figref{fig:fig4} presents clear and illustrative examples of the quality differences among views  downsampled to a resolution of $224^2$ from different rendering resolutions. Compared with the direct rendering of CPMF to $224^2$, our method eliminates most of the noise, enhances the quality of the rendered images, and thus further improves the feature representation ability. Additionally, for sparse 3D data, attention should be paid to noise introduced by regional data voids under high-resolution rendering.

\subsubsection{ Influence of Rendering Views Numbers }
The saliency of different defect types varies in depth images projected from different angles, so the number of projection views has a significant impact on the performance. To study the influence of $N_v$, this study conducts several sets of experiments using different $N_v$, $N_v \in \left \{  1, 3, 6, ... , 21 \right \}$. As \figref{fig:fig5} shows, the experimental results demonstrate moderate yet consistent improvements in both O-ROC and P-ROC metrics with increasing view quantity $N_v$. The most significant enhancement occurs when $N_v$ rises from 1 to 3, O-ROC peaks at around 15 views, while P-ROC reaches its highest value at approximately 21 views. Performance at $N_v$ =15 shows substantial improvement compared to $N_v$ =1. Notably, during the detection process, the MVR framework achieves superior detection performance with only three depth views as input. This characteristic validates MVR's ability to extract discriminative features from limited data while also highlighting the method's strong robustness and significant advantages in computational efficiency and resource utilization. Nevertheless, the observed fluctuations in experimental metrics may primarily be attributed to the selection of specific viewpoints containing low-quality features. These low-quality features lead to inconsistent feature representations across different perspectives, resulting in feature distribution mismatches within the latent space, thereby degrading the discriminative capabilities of anomaly detection models.

\begin{table}[htbp]
  \centering
  \caption{{The Performance Using Different Backbones on Real3D-AD~\cite{read3d}}.}
  \begin{tabular}{lccccc}
    \toprule
    \textbf{Arch.} & \textbf{Param.} & \textbf{MACs} & \textbf{ O-ROC\%} & \textbf{ P-ROC\%} \\
    
    \midrule
    ViT-Small & 37.4M & 26.3G & 88.1 & 94.3 \\
    ViT-Base & 148.0M & 104.7G & 89.3 & 95.3  \\
    \rowcolor{blue!8} 
    ViT-Large & 275.3M & 413.5G & \textbf{89.6} & \textbf{95.7} \\
    \bottomrule
  \end{tabular}
\end{table}

\subsubsection{ Influence of Backbones }
To evaluate the impact of backbone architectures, we conduct comparative analyses using ViT-L, ViT-B, and ViT-S with point-wise evaluation metrics (Table 2). Experimental results demonstrate significant performance disparities among the architectures, where ViT-L consistently surpasses both ViT-B and ViT-S. This empirical evidence indicates that scaling network capacity enhances the model's ability to learn generalized representations and capture subtle anomaly characteristics.

\section{Conclusions}
In this work, we propose the MVR framework, which innovatively integrates the 3D-oriented 2D feature extraction paradigm with geometric reconstruction constraints. This approach effectively addresses the over-reliance of traditional methods on local feature matching and strengthens the learning of global semantics. Experimental results validate the framework's effectiveness in complex industrial scenarios, establishing an highly efficient and robust new paradigm for 3D anomaly detection that holds significant value for practical applications.

\addtolength{\textheight}{-12cm}   % This command serves to balance the column lengths

%%%%%%%%%%%%%%%%%%%%%%%%%%%%%%%%%%%%%%%%%%%%%%%%%%%%%%%%%%%%%%%%%%%%%%%%%%%%%%%%

%%%%%%%%%%%%%%%%%%%%%%%%%%%%%%%%%%%%%%%%%%%%%%%%%%%%%%%%%%%%%%%%%%%%%%%%%%%%%%%%

%%%%%%%%%%%%%%%%%%%%%%%%%%%%%%%%%%%%%%%%%%%%%%%%%%%%%%%%%%%%%%%%%%%%%%%%%%%%%%%%

%%%%%%%%%%%%%%%%%%%%%%%%%%%%%%%%%%%%%%%%%%%%%%%%%%%%%%%%%%%%%%%%%%%%%%%%%%%%%%%%

{\small
% % \bibliographystyle{plain,unsrt}
% \bibliographystyle{abbrv}

% % \bibliography{bibtex/bib/IEEEabrv}
% \bibliography{ref}

% \bibliographystyle{plain,unsrt}
% \bibliographystyle{unsrt}
\bibliographystyle{ieeetr}
% \bibliography{bibtex/bib/IEEEabrv}
\bibliography{ref}

}

\end{document}